\documentclass{article}

\usepackage{arxiv}
\usepackage[utf8]{inputenc} 
\usepackage[T1]{fontenc}    
\usepackage{hyperref}       
\usepackage{url}            
\usepackage{booktabs}       
\usepackage{amsfonts}       
\usepackage{nicefrac}       
\usepackage{microtype}      
\usepackage{graphicx}       
\usepackage{subcaption}     
\usepackage{multirow}

\usepackage{float}
\title{Capsule GAN for Prostate MRI Super-Resolution}

\author{
  Mahdiyar Molahasani Majdabadi \\
  Department of Electrical \& Computer Engineering\\
  University of Saskatchewan\\
  \texttt{m.molahasani@usask.ca}\\
  \And
    Younhee Choi \\
  International Road Dynamics, Canada\\
  \texttt{younhee.choi@irdinc.com}\\
   \And
  S. Deivalakshmi \\
  Department of Electrical \& Computer Engineering\\
  National Institute of Technology, Trichy, India\\
  \texttt{deiva@nitt.edu}\\
  \And
  Seokbum Ko \\
  Department of Electrical \& Computer Engineering\\
  University of Saskatchewan\\
  \texttt{seokbum.ko@usask.ca}\\
}

\begin{document}
\maketitle

\begin{abstract}

Prostate cancer is a very common disease among adult men. One in seven Canadian men is diagnosed with this cancer in their lifetime. Super-Resolution (SR) can facilitate early diagnosis and potentially save many lives. In this paper, a robust and accurate model is proposed for prostate MRI SR. The model is trained on the Prostate-Diagnosis and PROSTATEx datasets. The proposed model outperformed the state-of-the-art prostate SR model in all similarity metrics with notable margins. A new task-specific similarity assessment is introduced as well. A classifier is trained for severe cancer detection and the drop in the accuracy of this model when dealing with super-resolved images is used for evaluating the ability of medical detail reconstruction of the SR models. The proposed SR model is a step towards an efficient and accurate general medical SR platform.  

\end{abstract}


\section{Introduction}
Cancer is one of the most common reasons for death worldwide. Effective treatment plans can significantly increase the survival rate of the patients and increase their life expectancy and quality. One of the most important factors in the success of a treatment plan is early diagnosis \cite{marini2006epigenetic}. Because of the nature of cancer, the earlier the disease is diagnosed, the more successful the treatment can be. One of the most common methods for cancer detection, especially in breast and prostate cancer, is medical imaging \cite{islam2013survey,akbari2012hyperspectral}. Using various technologies such as MRI and ultrasound, an image of internal tissue is taken and the doctor checks the image for any abnormality and signs of cancer. In the early stage, these abnormalities are usually very small so because of the low resolution of the scans and high noise level, it is impossible to detect them. 

The second most common cancer in male individuals is prostate cancer \cite{brenner2020projected}. One of the most effective approaches for early diagnosis and treatment planning in this cancer is MRI \cite{kasivisvanathan2018mri}. In order to address the aforementioned challenges for cancer detection, especially in prostate cancer, AI can help the diagnosis process by increasing the resolution of the scan and improve the quality of the image \cite{wang2018esrgan,amaranageswarao2020residual}. Thus, more detailed scans can be provided using the same equipment and the disease can be diagnosed at an earlier stage.  Super-Resolution (SR) is an effective way to increase the quality of a scan and reduce the noise level. Motivated by these facts, we have implemented a deep learning model for prostate MRI SR in order to facilitate early diagnosis and help save lives. Although there are many researches on SR for medical scans, currently, there are not many works on high-scale SR in this domain \cite{amaranageswarao2020wavelet}. The main concern is whether the medical details essential for the diagnosis will be preserved in the process.  

SR in medical imaging has been around for a while. The early works used image processing algorithms for medical SR. Rousseau \textit{et al.} employed anatomical inter-modality priors obtained from a particular reference sample \cite{rousseau2010non}. Peeters \textit{et al.} interpolated slice-shifted images in order to enhance signal to noise ratio \cite{peeters2004use}. Moreover, the effective slice thickness was decreased as well.  

More recently, machine learning algorithms have become more popular for medical SR. A 2D convolutional network has been used by Yang \textit{et al.} and Park \textit{et al.} \cite{yang2016super,park2018computed}. 3D SR was addressed by Chaudhari \textit{et al.} and Chen \textit{et al.} using 3D architectures \cite{chaudhari2018super,chen2018brain}. These machine learning approaches surpassed traditional image processing algorithms in both visual quality and computational complexity.

Recently, powerful DL models such as GANs have improved the performance of the SR models significantly and made high-scale medical SR possible \cite{goodfellow2014generative,brock2018large}. Hence, it has been widely used for the SR task due to its natural-looking outputs. Shi \textit{et al.} uses GAN for single image super-resolution \cite{shi2018super}. Perceptual loss function, which consists of adversarial loss and content loss, is used for training the GAN for the SR task. It is demonstrated that  GAN performs the best for medical SR among all machine learning algorithms \cite{sood2018application}. A GAN using 3D convolutional layers is proposed by Li \textit{et al.} for slice thickness reduction in MRI. SRGAN, as one of the first GAN models for SR, was utilized by Sood \textit{et al.} for normal and anisotropic SR in prostate MRI \cite{li2017reconstruction}. Their findings suggest that the visual quality of the super-resolved images obtained by SRGAN is notably better than other approaches.        

In \cite{majdabadi2020msg}, we have proposed Multi-Scale Gradient Capsule GAN (MSG-CapsGAN) network for face SR. This model is the first MSG-GAN used for SR. Moreover, this is the first SR model that benefits from Capsule Network (CapsNet) for SR. In \cite{MolahasaniMajdabadi2020}, we have improved the architecture of MSG-CapsGAN and used patch based architecture alongside with residual learning. Hence, the proposed model surpassed the state-of-the-art face SR model. Since unlike the related face SR models, the proposed model is not using any attribute domain data, it can be considered as a general SR system. Motivated by the promising performance of the proposed model, this architecture is modified and employed to address early prostate cancer detection by performing SR on MRI scans.

In this paper, a powerful deep learning architecture for prostate MRI SR is proposed. This model is benefiting from MSG-GAN and CapsNet in its architecture. Moreover, CheXNet is embedded in the model as the feature extractor. A novel task-specific similarity metric is also introduced which evaluates the similarity images from the deep learning perspective. The results demonstrates that the proposed model outperforms the state-of-the-art prostate MRI 
SR model in all similarity metrics. The contributions of this paper are summarized as follows:

\begin{enumerate}
    \item This is the first time that MSG-GAN and CapsNet is employed for medical SR. MSG-GAN improves the quality of the output in high-scale SR and CapsNet enhanced the robust of the model and the ability to generalize.  
    
    \item Relying on the MSG-GAN architecture and benefiting from CapsNet in the discriminator, the proposed model outperforms the state-of-the-art systems in terms of PSNR, Structural SIMilarity (SSIM), Multi-Scale Structural SIMilarity (MS-SSIM), and Task-Specific Similarity Assessment (TSSA) with 5.5\%, 27.6\%, 21\%, and 7.3\%, respectively.
    
    \item The proposed model is more compact and has 42\% less number of trainable parameters than the state-of-the-art prostate MRI SR model.

\end{enumerate}

The proposed model is a step towards a general and accurate high-scale SR system for medical scans that can facilitate early diagnosis and save many lives.

The paper is structured as follows. After the introduction, prostate cancer and its main diagnosis approaches are reviewed in Chapter 2. Then, in Chapter 3, the concept of MSG-GAN is explained. Chapter 4 presents the proposed model architecture with details. The dataset and preprocessing algorithms are addressed in Chapter 5. The results are explained and discussed in Chapter 6, followed by conclusion in Chapter 7.

\section{Prostate Cancer}
\begin{figure}[H]
	\centerline{\includegraphics[width=0.37\linewidth ]{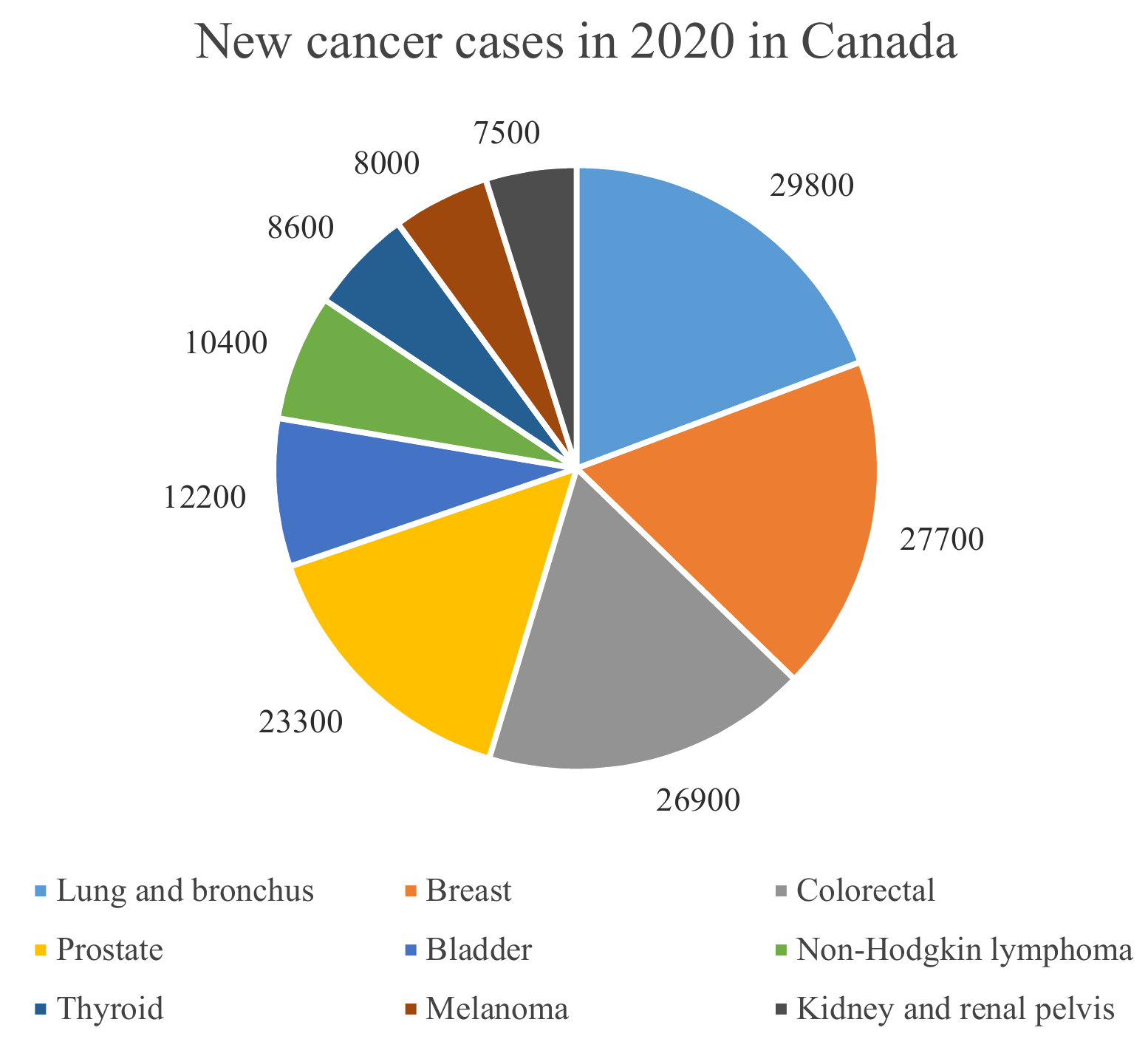}}
	\caption{The distribution of the new cancer cases in Canada in 2020 \cite{brenner2020projected}.}
	\label{cancer}
\end{figure}

In 2020, the fourth most common cancer in Canada was prostate cancer, as illustrated in Fig. \ref{cancer}.

This cancer is identified as the most prevalent non-cutaneous cancer in men. It is also the second-cancer-related cause of death in male individuals. Worldwide, 1 in every 9 men will be diagnosed with prostate cancer through their lifetime \cite{reda2018deep}.

There are two major approaches for monitoring the symptom-free patient for possible prostate cancer:

\begin{itemize}
	\item \textbf{Digital rectal exam (DRE)} is a test in which the doctor examines the rectum for identifying any abnormality in the shape, size, or texture of the prostate. 
	
	\item \textbf{Prostate-specific antigen (PSA) test} is a blood test used for prostate cancer detection. In this test, the density of PSA in the patient's blood is analyzed. A high level of PSA can be an indication of infection, inflammation, or cancer in the prostate. 
	
\end{itemize}

If any abnormality is detected in the aforementioned tests, the following approaches are used for cancer detection:

\begin{itemize}
	\item \textbf{Ultrasound:} A prob is used to take an image of the prostate using ultrasound waves.
	
	\item \textbf{Magnetic resonance imaging (MRI):} More detailed images of the prostate can be acquired using MRI. This image can be used by the doctor not only for cancer detection but also for treatment planning. 
	
	\item \textbf{Collecting a sample of prostate tissue:} A tissue sample from the prostate is collected through a prostate biopsy. The presence of the cancer cells is investigated in the lab for disease confirmation.  
	
\end{itemize}

\begin{figure}[H]
	\centerline{\includegraphics[width=\linewidth ]{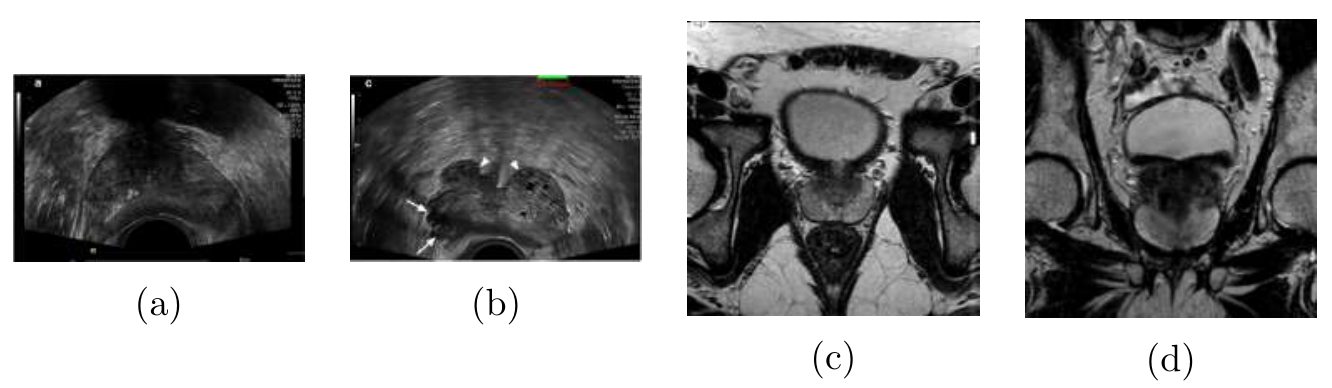}}
	\caption{Samples of ultrasound images of (a) healthy prostate \cite{liau2019prostate}, (b) prostate with cancer \cite{liau2019prostate}, and the MRI of (c) healthy prostate \cite{us2021}, (d) prostate with cancer \cite{prostatex}.}
	\label{usmri}
\end{figure}

Fig. \ref{usmri} depicts two different types of prostate diagnosis imaging methods. 

MRI can provide very high-quality images of the prostate. As a result, it is a very helpful way to detect the disease. However, the quality of scans is not always adequate for early-stage cancer detection, especially in cheaper scanning devices. The commonness of prostate cancer and the importance of early diagnosis, alongside the ability of MRI in acquiring good images, motivates us to address MRI SR in prostate cancer.

\section{MSG-GAN}

\begin{figure}[H]
	\centerline{\includegraphics[width=0.3\linewidth ]{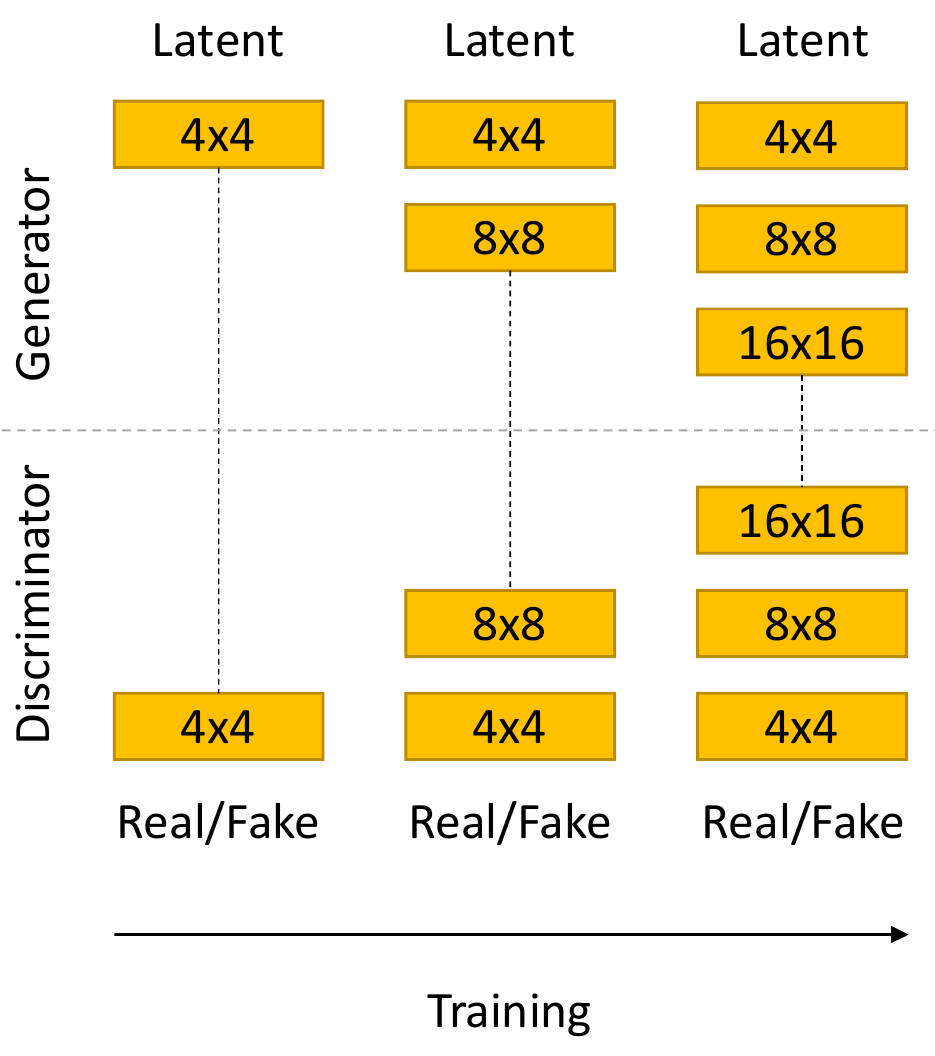}}
	\caption{The architecture of progressive GAN \cite{karras2017progressive}.}
	\label{pganarch}
\end{figure}

One major challenge in generating a high-resolution image using GAN is the gradient problem. This problem occurs when the distribution of the generated images have not enough overlap with the distribution of the training set. To overcome this issue, a layer-wise solution is proposed \cite{karras2017progressive}. Training GANs progressively is an effective training approach to generate HR images. Fig. \ref{pganarch} displays the architecture of progressive GAN.

\begin{figure}[H]
	\centerline{\includegraphics[width=0.25\linewidth ]{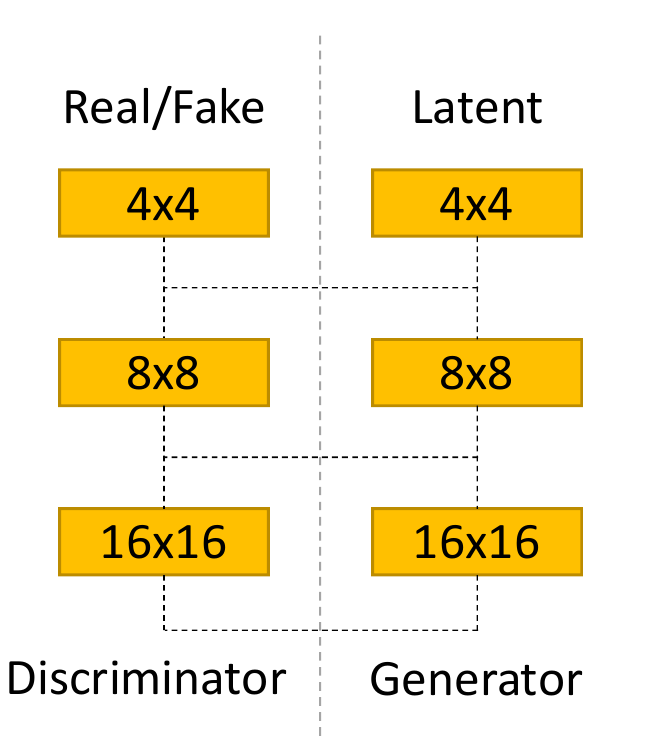}}
	\caption{The architecture of MSG-GAN \cite{karnewar2020msg}.}
	\label{msggan}
\end{figure}

First, both generator and discriminator have a very simple architecture and the model is trained to create LR fake images. In the next step, a new layer is gradually added to both models which enable the GAN to double the output resolution. This process continues to the point that the output size reaches the desired dimension. Since the first layers are trained at first, the gradient problem is resolved. However, this architecture and training paradigm is so complex. A simpler yet effective solution called MSG-GAN is proposed by Karnewar \textit{et al.} \cite{karnewar2020msg}. This alternative solution attempts to solve the gradient problem using a fixed model, i.e., non-progressive, with new paths for the gradient. Fig. \ref{msggan} depicts the architecture of MSG-GAN.   

Unlike progressive GAN, the architecture of MSG-GAN does not change through time. The intermediate connections create new paths for the gradient. In the progressive model, to make sure that the distribution of the fake and real samples in low-scale SR is the same, only the first layer of the models was present in the training. However, in MSG-GAN, the connection between the first layer of the generator and the first layer of the discriminator helps the model to overcome the gradient problem. 

In medical SR, reconstructing all medical details useful for diagnosis is crucial. As a result, overcoming the gradient problem in GAN-based SR models is necessary. Motivated by the performance of MSG-GAN, and the flexibility and stability of this architecture, this model is used in this research to form MSG-CapsGAN for the SR problem. More details on the architecture of the proposed GAN for MRI SR are provided in the next Chapter.

\section{Proposed Model Architecture}
\begin{figure}[H]
	\centerline{\includegraphics[width=0.7\linewidth ]{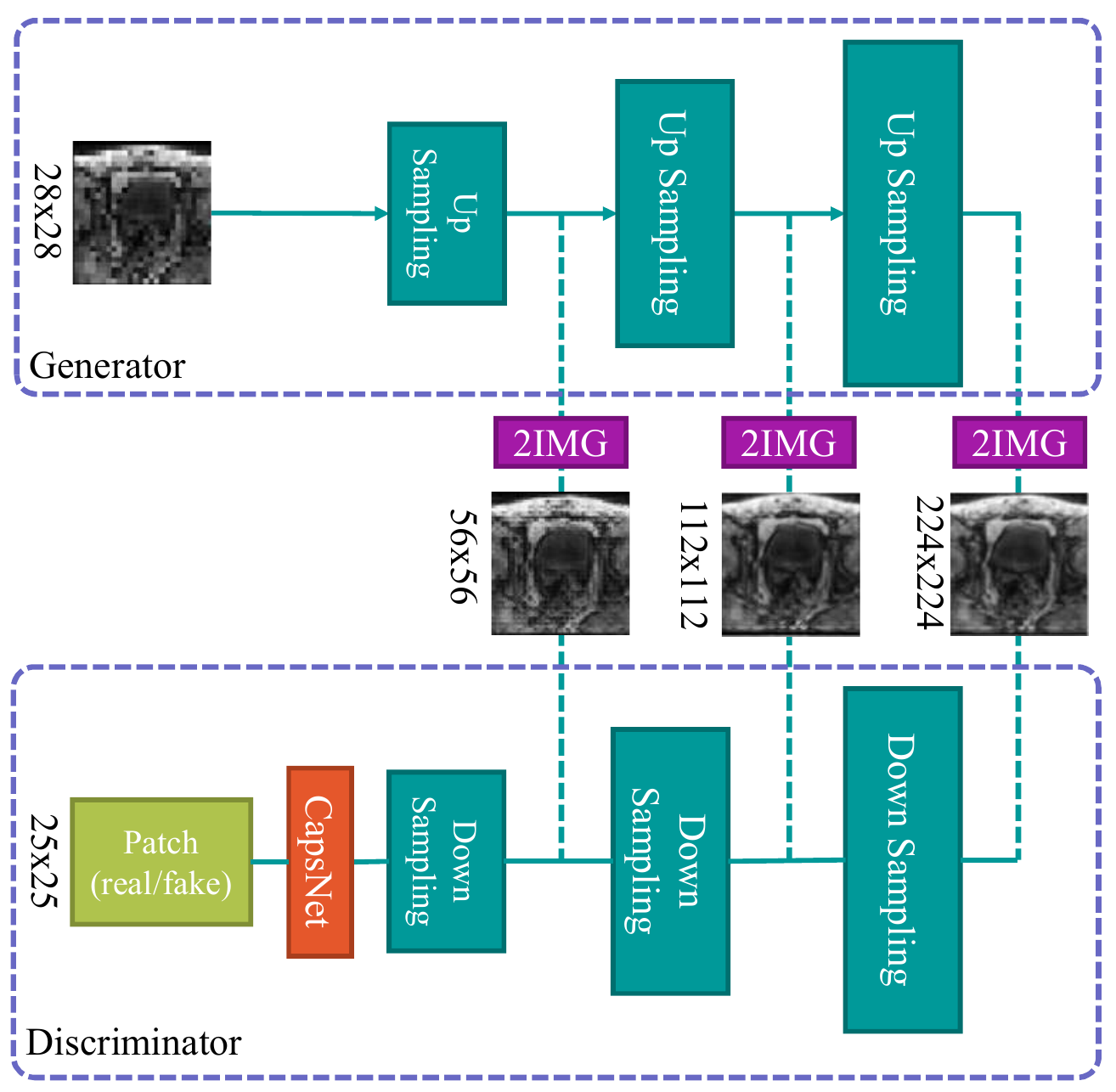}}
	\caption{The architecture of the model for prostate MRI SR.}
	\label{all_arch_ill}
\end{figure}

The architecture used in this study is based on the model proposed in our previous work, originally for face SR \cite{MolahasaniMajdabadi2020}. This model is the first Multi-scale Gradient Capsule GAN and outperforms state-of-the-art face SR systems in all similarity metrics. This network is modified and used for addressing the high-scale prostate MRI SR. The architecture of the proposed model is shown in Fig. \ref{all_arch_ill}. 

The proposed model consists of two networks. The generator reconstructs the HR image from LR input. After each up-sampling unit, the output is passed to $2IMG$ layer. This layer is a convolutional layer with one filter that creates a black and white output image. These images are shared with the discriminator. The discriminator uses these images to determine whether the input images belong to the dataset or they are super-resolved samples. The outputs of down-sampling layers are passed to the CapsNet.

\begin{figure}[H]
	\centerline{\includegraphics[width=\linewidth ]{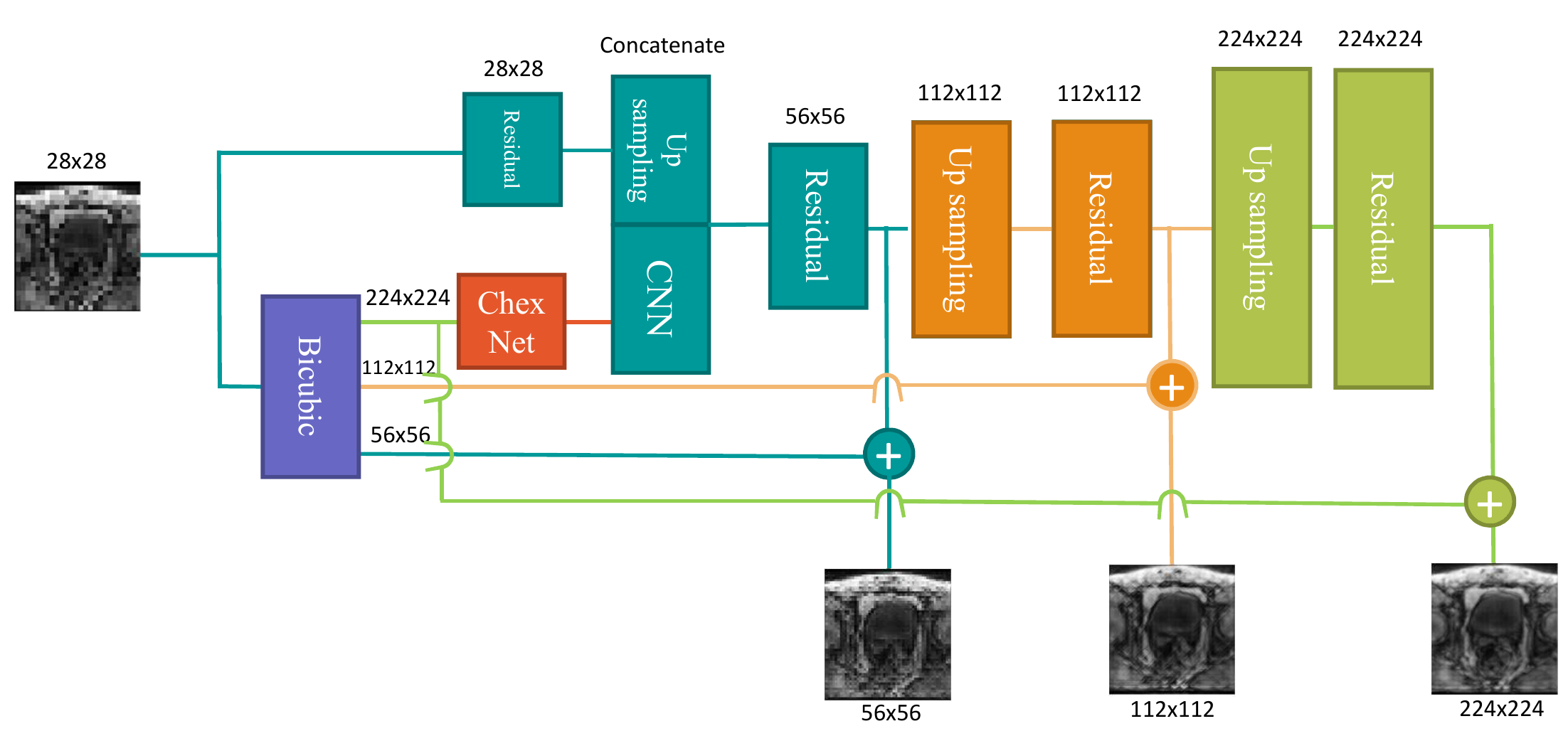}}
	\caption{The architecture of the proposed generator for prostate MRI SR.}
	\label{parch}
\end{figure}

CapsNet is a deep learning architecture consisting of capsules. Each capsule is a group of neurons representing a vector. The length of this vector corresponds to the probability of a presence of a feature in the previous layer. The primacy of the CapsNet over previous architecture is this model is learning the relationship between the features as well. As a result, it can improve the training time and the robustness of the model. Motivated by these advantages of CapsNet, this network is employed in the proposed architecture as the classifier. The utilized CapsNet is a two layer capsule network with 8 capsules with 32 neurons in the first layer and 10 capsules with 16 neurons in the second layer. Finally, a fully connected layer with 625 neurons creates the $25 \times 25$ patch as the output. Each value in this matrix ranges from 0 to 1 representing fake or real, respectively. The more detailed architecture of the generator is depicted in Fig. \ref{parch}.

\begin{figure}[H]
    \centering
    \begin{subfigure}{0.45\linewidth}
        \centering
        \includegraphics[width=\linewidth]{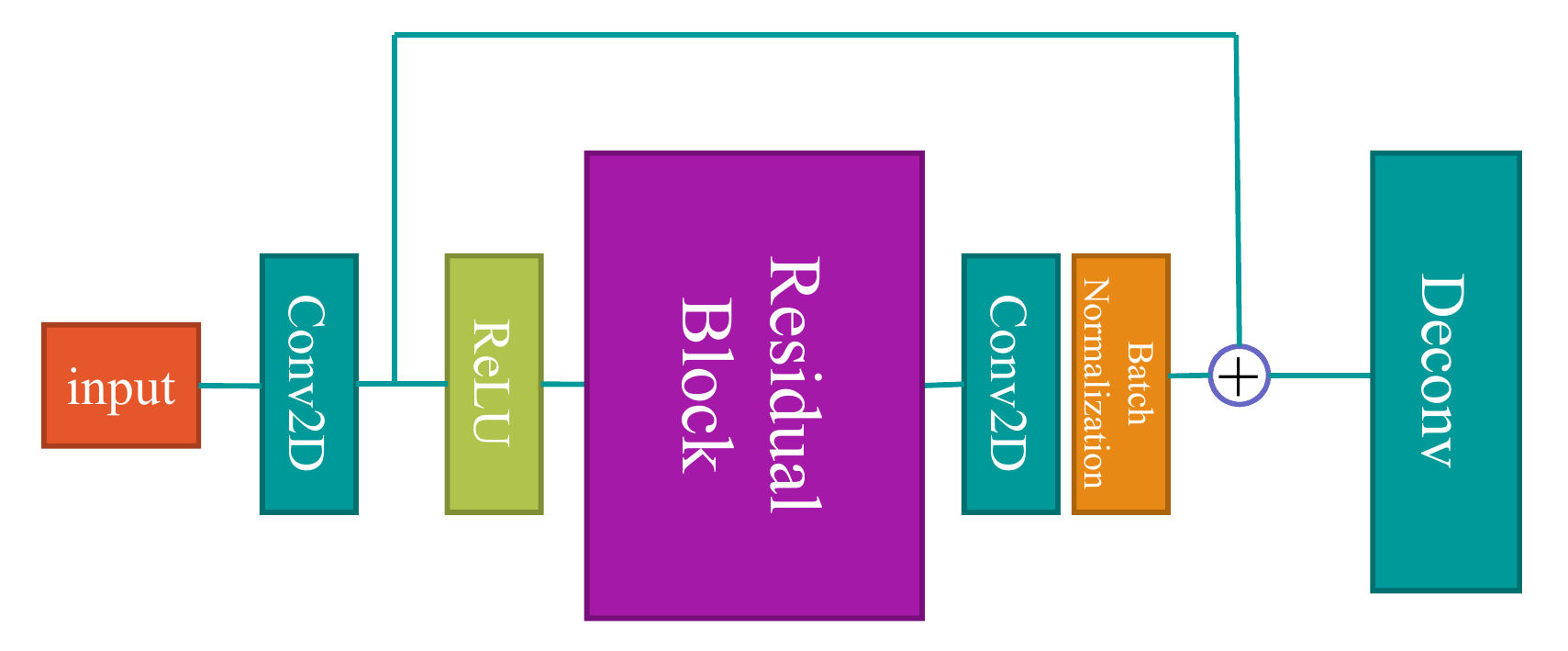}
        \caption{}
        \label{fig:test_uesb}
    \end{subfigure}
    \begin{subfigure}{0.45\linewidth}
        \centering
        \includegraphics[width=0.8\linewidth]{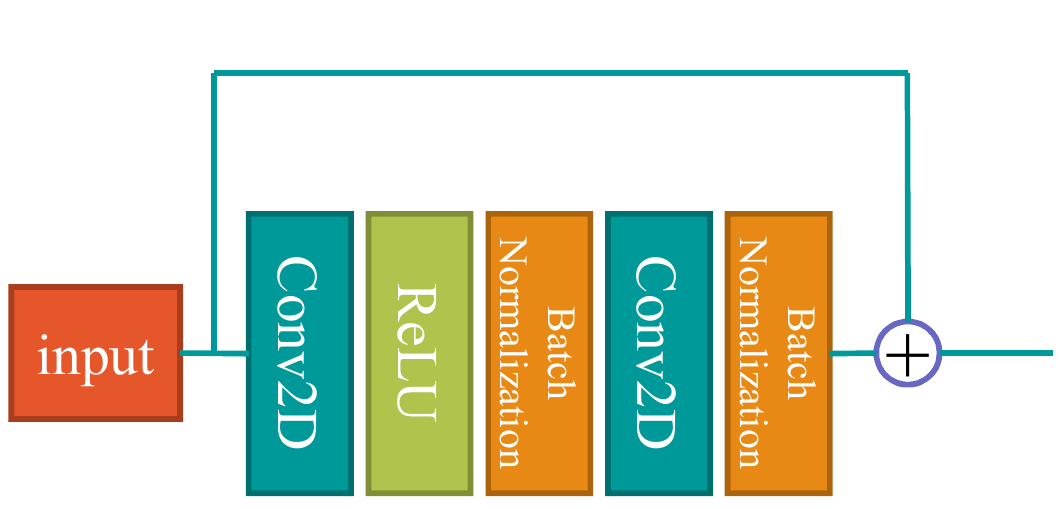}
        \caption{}
        \label{fig:test_uesb_mask}
    \end{subfigure}
\caption{The architecture of (a) up-sampling unit and (b) the residual block.}
\label{Resi}
\end{figure}

\noindent LR image is up-sampled with different scales with bicubic interpolation for residual learning and also feature extraction using ChexNet. The architecture of the up-sampleing unit and residual block used in Fig. \ref{parch} is illustrated in Fig. \ref{Resi}. 

The up-sampling and Residual block are benefiting from bypass connections. These connection are solving the vanishing gradient problem. As a result, significant improvement in the performance of the network can be witnessed \cite{he2016deep}. The extracted features are concatenated by a learnable CNN feature map and passed to a residual block. Each up-sampling block is followed by a residual block and the output of this block creates high-frequency details of the image in each scale. This high-frequency information is then added to a bicubic interpolation of the input image. 

Unlike the face SR problem, the size of the HR image in this study is $224\times 224$. Moreover, each MRI slice is a black and white image with one channel. Another significant difference between this architecture and the architecture in \cite{MolahasaniMajdabadi2020} is the feature extractor. In MSG-CapsGAN, VGG19 is used in the heart of the model. This model extracts useful features from the interpolated image and SR system is reconstructing the image using these informative extracted features. However, MRI images are completely different from the samples in the ImageNet dataset \cite{deng2009imagenet}. Each slice of MRI scan is a one-channel gray scale image while ImageNet samples are three-channel RGB images. In \cite{haghanifar2020paxnet}, the features extracted from a tooth radiography by Inception Net\cite{szegedy2015going}, a CNN trained on ImageNet dataset, and CheXNet is compared. It is demonstrated that CheXNet can extract more complex and informative features in comparison with Inception Net while Inception Net can mainly detect edges. Hence, embedding this model in the generator is not useful. Inspired by our previous work on caries detection on segmented teeth, in this model, VGG is substituted with CheXNet \cite{haghanifar2020paxnet,haghanifar2020automated}.

\begin{figure}[H]
	\centerline{\includegraphics[width=\linewidth ]{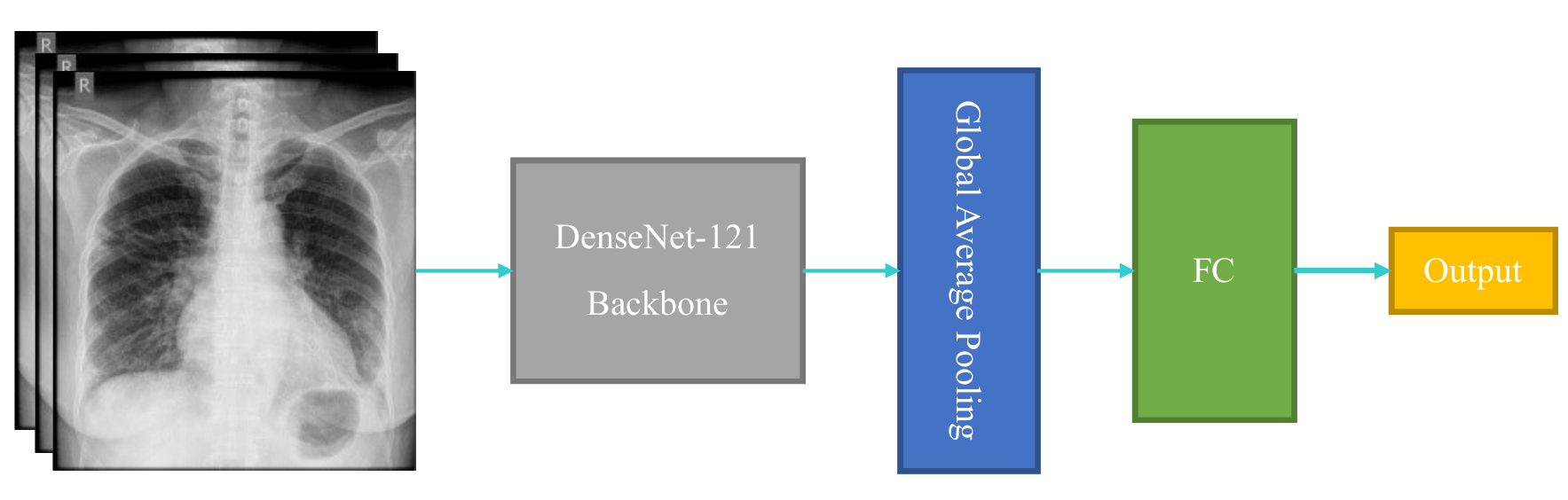}}
	\caption{The architecture of CheXNet \cite{rajpurkar2017chexnet}.}
	\label{chex}
\end{figure}

CheXNet is a deep learning architecture trained for lung disease classification. The architecture of CheXNet is based on DensNet as presented in Fig. \ref{chex}.

\begin{figure}[H]
	\centerline{\includegraphics[width=\linewidth ]{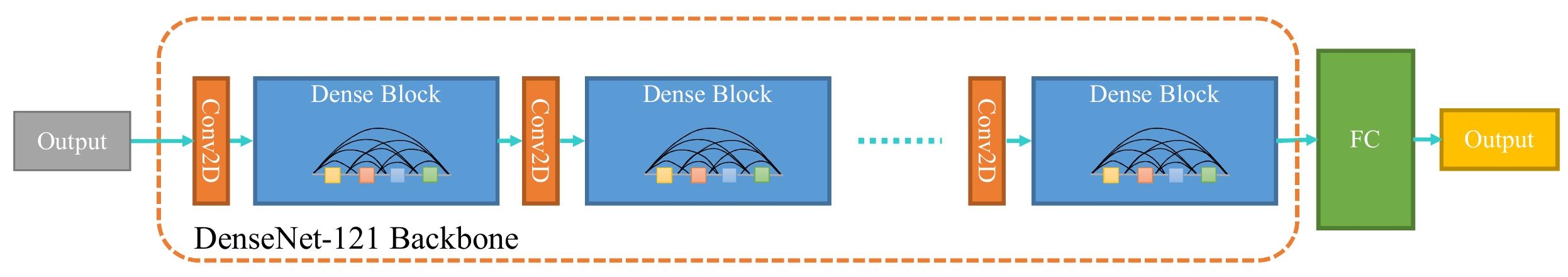}}
	\caption{The architecture of DenseNet-121 \cite{rajpurkar2017chexnet}.}
	\label{dense}
\end{figure}

The input is passed to DenseNet-121, followed by a global average pooling layer. Then, a Fully Connected (FC) generates the output. The architecture of DensNet is shown in Fig. \ref{dense}.

Each dense block has four layers and the output of each layer is connected to each layer after itself.

CheXNet has been trained on frontal chest x-ray images from  CXR datasets. CXR dataset is a large publicly available dataset of 14 different chest diseases. The frontal chest x-ray is indeed different from prostate MRI, however, the basic features extracted by the first layers of CheXNet are very informative. We have demonstrated that these features are useful even when they are applied to other radiology scans \cite{haghanifar2020paxnet}.

\section{Dataset and Preprocessing}
\begin{figure}[H]
	\centerline{\includegraphics[width=\linewidth ]{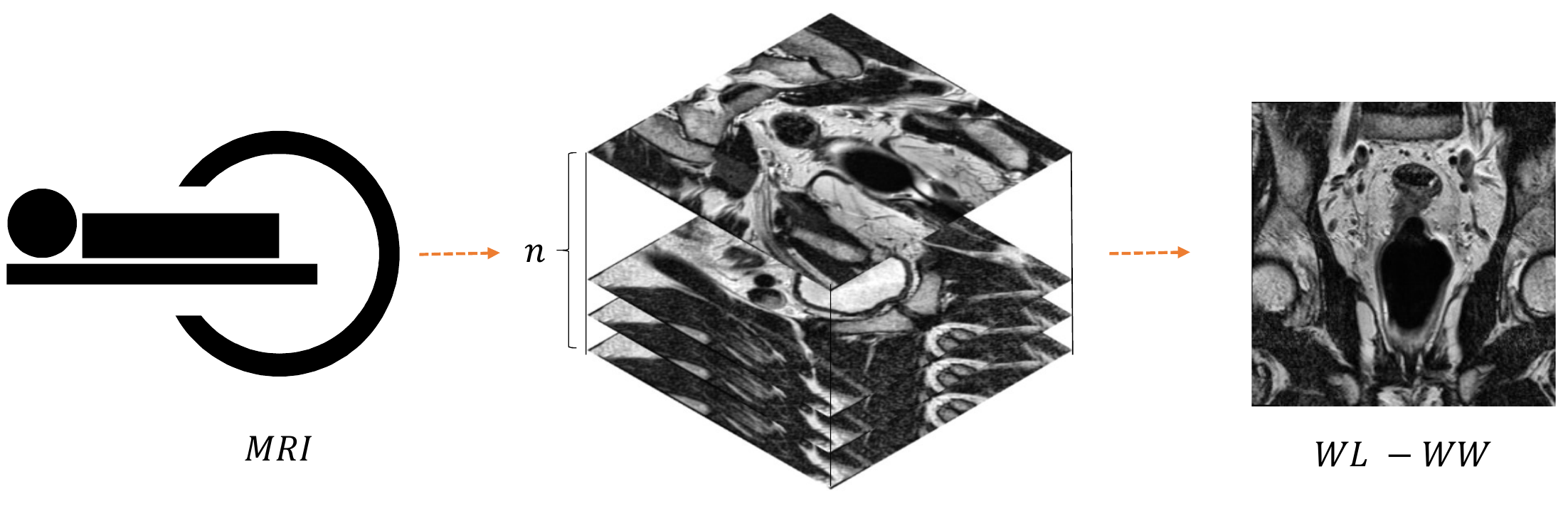}}
	\caption{The process of obtaining dataset.}
	\label{mri_step}
\end{figure}

Prostate-Diagnosis and PROSTATEx dataset are used in this work \cite{prostatex,PROSTATE-DIAGNOSIS}. These datasets are publicly available on Cancer Imaging Archive  \footnote{https://www.cancerimagingarchive.net/}. These two datasets contain multi-slice MRIs of patients all diagnosed with prostate cancer, as illustrated in Fig \ref{mri_step}.

\begin{figure}[H]
	\centerline{\includegraphics[width=\linewidth ]{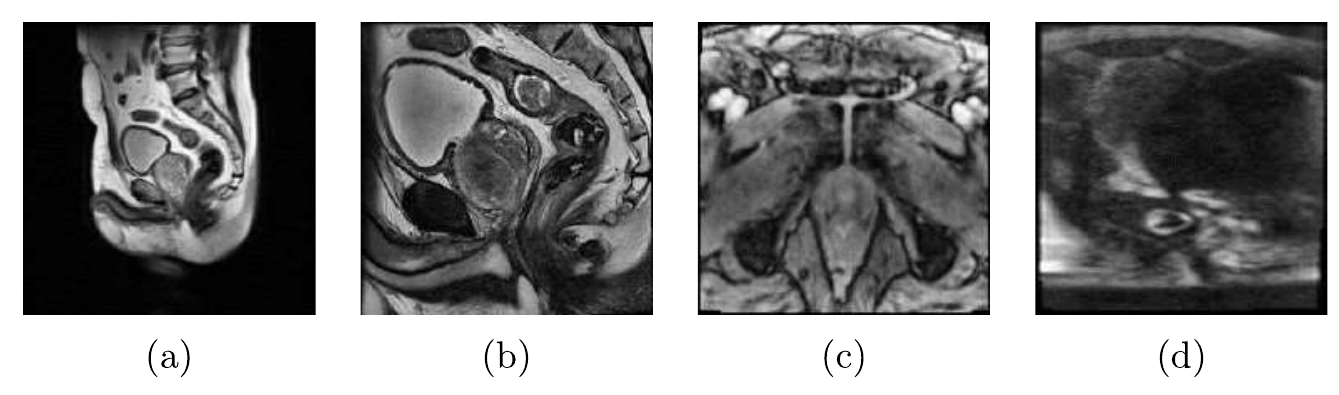}}
	\caption{Samples from PROSTATEx and Prostate-Diagnosis dataset (a) wide sagittal, (b) sagittal, (c) axial, and (d) low-quality axial.}
	\label{views}
\end{figure}

Where $n$ is the number of slices in each MRI. Each slice is a 16-bit black and white DICOM image. $WL$ and $WW$ are window level and window width, respectively. By changing these two parameters, the brightness and the contrast of the image can be adjusted.   
The data is obtained from 329 patients and each patient has several scans and each scan consists of many slices. The total number of images is 329k. The scans fall into three categories of coronal, sagittal, and axial with different qualities. Some samples from the dataset are exhibited in Fig. \ref{views}.

For the SR task, the data of 320 patients are used for training and the rest 9 are used for model performance evaluation. 

\begin{figure}[H]
	\centerline{\includegraphics[width=0.9\linewidth ]{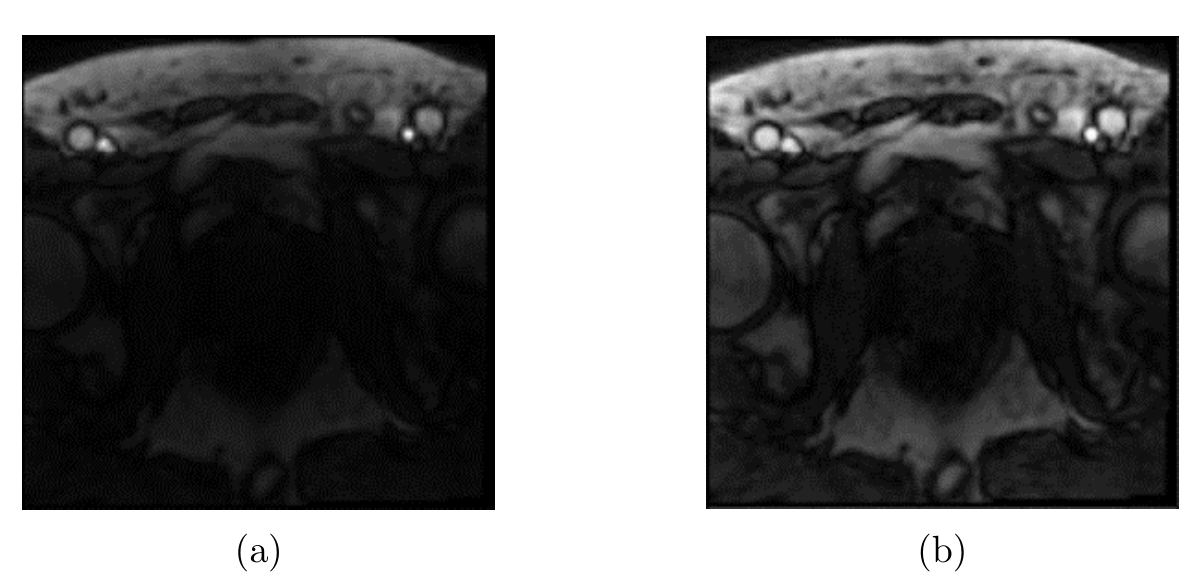}}
	\caption{Sample image from the dataset (a) without CLAHE and (b) after applying CLAHE.}
	\label{sample_data}
\end{figure}

All the samples in the dataset are in DICOM format so the first step of data preparation is reading DICOM images and resizing them to $224\times 224$. Then, Contrast Limited Adaptive Histogram Equalization (CLAHE) is applied to each image. CLAHE is one of the most popular and powerful image enhancement algorithms \cite{pizer1987adaptive}. Fig. \ref{sample_data} exhibits a sample from the dataset before and after applying CLAHE. 

\noindent By equalizing the histogram in each region of the image, more details become visible. This preprocessing step can help the model to benefit from these details for HR image reconstruction. All the images are saved in PNG format. 

In each iteration of the training, a batch of images is loaded into the memory. Then, the images are flipped horizontally with the probability of 50$\%$. The image is down-sampled with scales of 2, 4, and 8. Finally, the format of the images changes from uint8 to float. 

PROSTATEx dataset contains metadata for each patient. One of the columns in the metadata is called ClinSig. ClinSig represents whether the scan is a clinically significant finding or not. When biopsy Gleason Score is 7 or higher, this identifier will be True. This label is used as the target label for training the classifier. 4K random images are selected from the dataset with their labels. 80$\%$ are used for training and the rest 20$\%$ for testing. A problem here is dataset imbalance. Only $20\%$ of the images are labeled as one. So in order to overcome this issue, an equal number of samples of each class is selected randomly from the dataset, instead of just a random choice regardless of the label.  

\section{Results and discussions}
The model is trained with a batch size of 32. In order to avoid mode collapse and prevent the discriminator from outperforming the generator, the discriminator is trained on real samples and fake images with a batch size of 8. Other training parameters are the same as in Table 3.1.

\subsection{Similarity Assessment}
To evaluate the performance of the model, the following similarity assessment metrics have been used:
\begin{itemize}
	\item Peak to Signal Noise Ratio (PSNR)
	\item Structural SIMilarity (SSIM)
	\item Multi-Scale Structural SIMilarity (MS-SSIM)
\end{itemize} 

\noindent These metrics evaluate the similarity of the super-resolved image with the ground truth. While PSNR is a pixel-wise metric, SSIM and MS-SSIM are evaluating the similarity of the distribution of the pixel values. It has been shown that these metrics fail to reflect the perceptual similarity of the output \cite{sood2018application}. As a result, Sood \textit{et al.} employed Mean Opinion Score (MOS) \cite{sood2018application}.

There are some important drawbacks to using MOS. First, it is extremely unlikely to reproduce the results, accurately. Hence, a comparison between models using MOS is almost impossible. Second, individuals participating in the process might be biased based on the questioner. More importantly, regarding the current performance of the DL models, the output of the SR model is usually passed to a DL-based classifier for automated diagnosis. 

Regarding the problems with each aforementioned metric, a Task-Specific Similarity Assessment (TSSA) metric is proposed. TSSA evaluates the similarity of super-resolved images and the ground truth from the perspective of a DL model by investigating the impact of the SR process on the performance of the DL model. TSSA can be formulated as:

\begin{equation} \label{fsim_3}
	TSSA = \frac{L_{gt}}{L_{SR}}
\end{equation}

\begin{figure}[H]
	\centerline{\includegraphics[width=0.7\linewidth ]{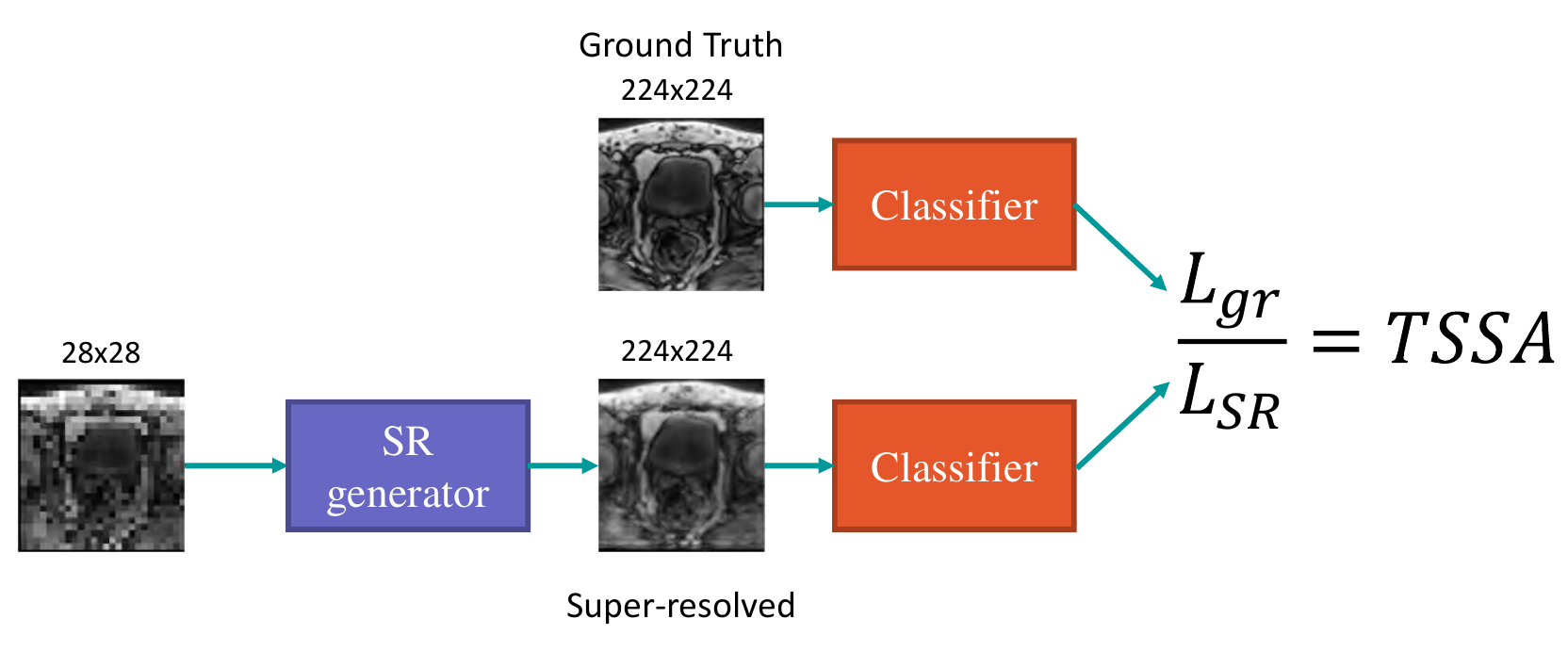}}
	\caption{TSSA calculation for a SR model.}
	\label{TSSA}
\end{figure}

\noindent Where $L_{gt}$ and $L_{SR}$ are the test loss of the classifier using ground truth and super-resolved images, respectively. Various functions can be used as $L$ for computing TSSA. Here the accuracy score is chosen to reflect the performance of the classifier after applying SR. TSSA is usually ranged from 0 to 1. When SR has no impact on the performance of the model, TSSA will be equal to 1. The more negative impact SR has on the performance of the classifier, the closer TSSA gets to 0. In a rare condition that SR improves the accuracy of the model, TSSA will become greater than 1. In this problem, a classifier for ClinSig classification is used. This CNN is trained with real $224\times 224$ prostate scans. Then the super-resolved images are used for evaluating the TSSA, as illustrated in Fig. \ref{TSSA}.

\subsection{Experimental Results}

\begin{figure}[H]
	\centerline{\includegraphics[width=\linewidth ]{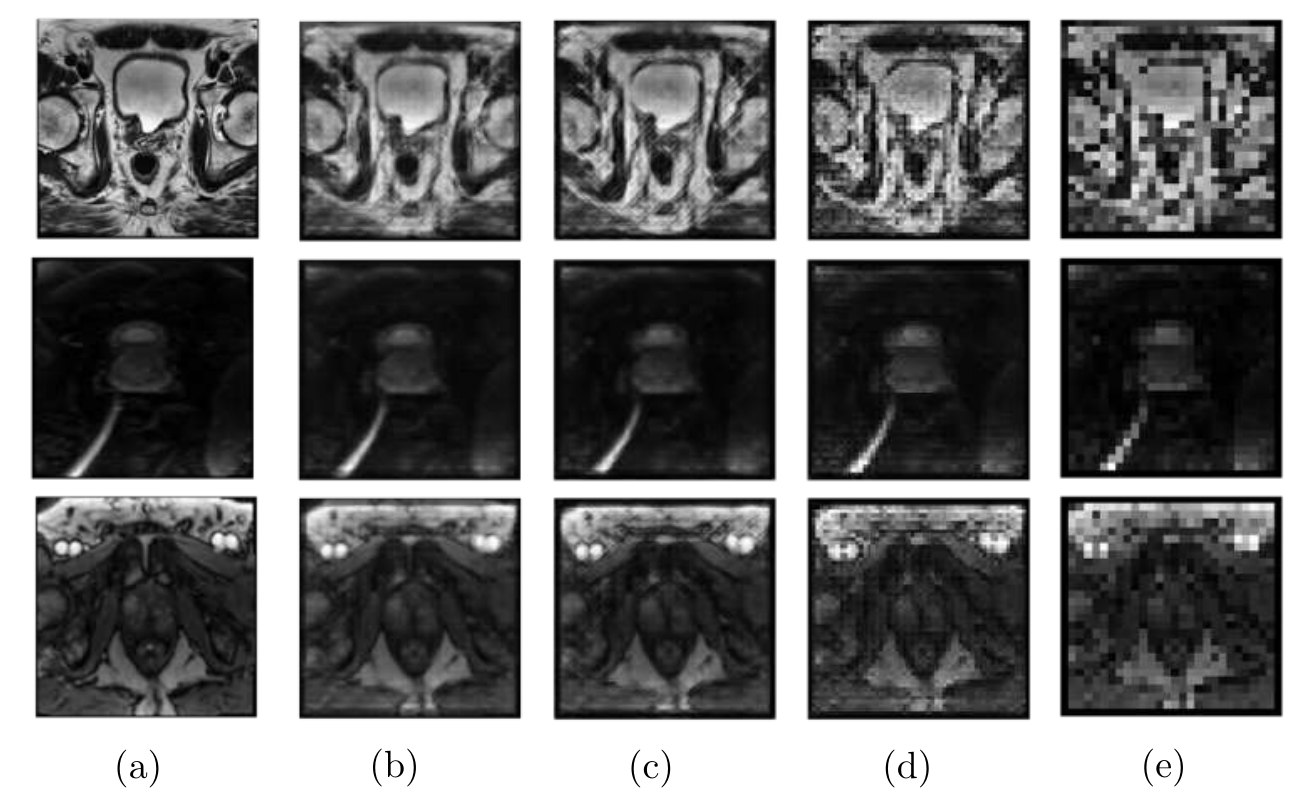}}
	\caption{(a) Ground truth, SR out put of the proposed model with the scale of (b) $8\times$, (c) $4\times$, (d) $2\times$, and (e) LR.}
	\label{stages}
\end{figure}

After training the model, the test set is used for performance evaluation. Fig. \ref{stages} shows the output of the model in different stages. 

The LR image is up-sampled for different scales. Each output is connected to the discriminator. As a result, the discriminator is classifying fake and real images based on not only the $\times 8$ images but also other scales as well.

\begin{figure}[!htbp]
	\centerline{\includegraphics[width=\linewidth ]{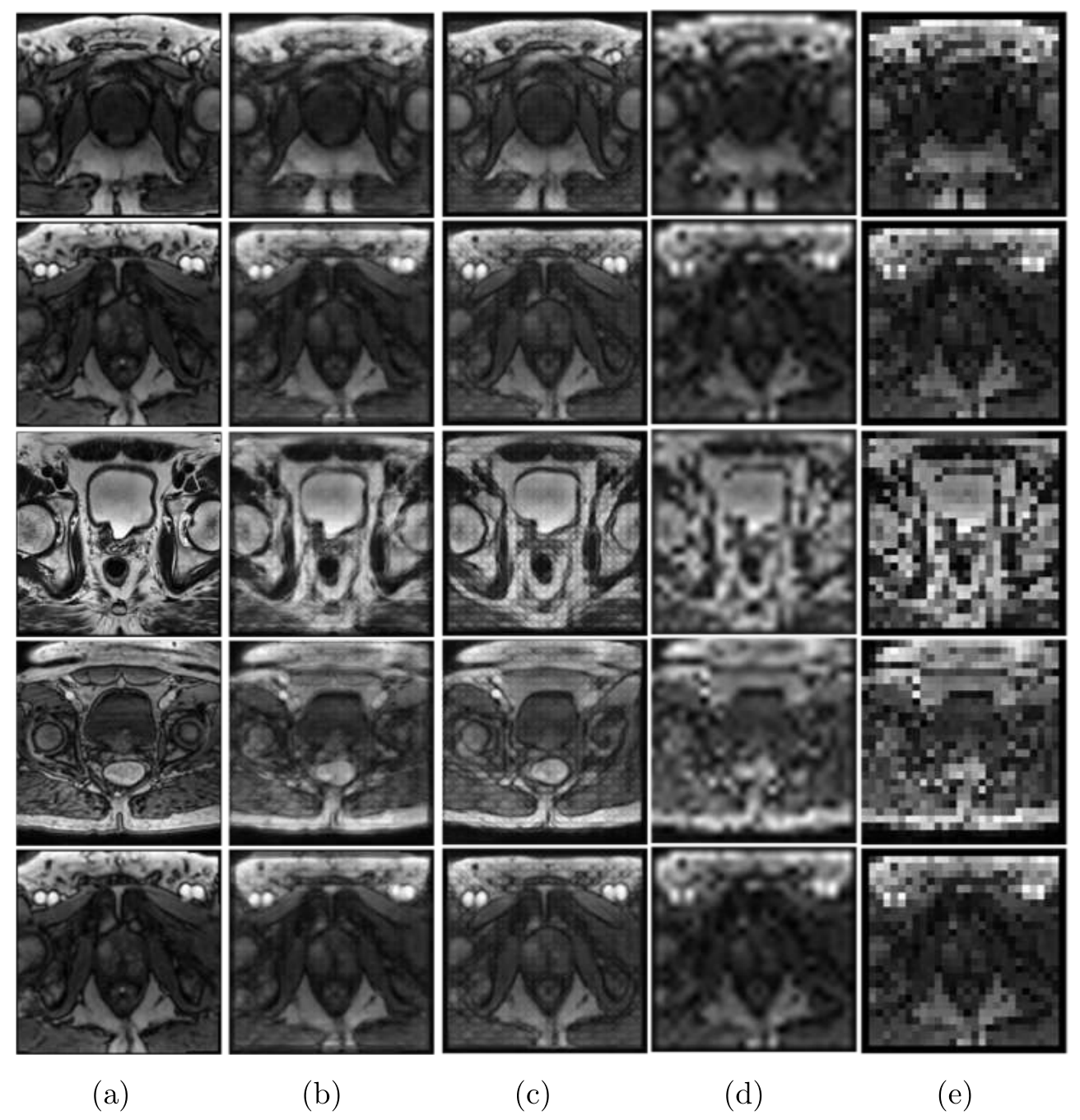}}
	\caption{(a)Ground truth, (b)proposed model, (c)SRGAN, (d)bicubic, and (e)LR.}
	\label{compare_sota}
\end{figure}

The proposed model is compared with the state-of-the-art approach which is an SRGAN \cite{sood2019anisotropic}. Fig. \ref{compare_sota} displays a visual comparison between the outputs of the models.

To compare the models quantitatively, PSNR, SSIM, and MS-SSIM are evaluated and presented in Table \ref{compare_metric}.

\begin{table}[!htbp]
	\caption{Comparison of the performance of different prostate SR models for $8 \times$ SR.}
	
	\begin{center}
		\begin{tabular}{|c|c|c|c|}
			\hline
			\textbf{Model} & \textbf{PSNR} & \textbf{SSIM} & \textbf{MS-SSIM} \\ \hline
			Bicubic        & 17.92         & 0.50          & 0.69             \\ \hline
			SRGAN \cite{sood2019anisotropic}          & 18.73         & 0.47          & 0.64             \\ \hline
			Proposed model & \textbf{19.77}         & \textbf{0.60}          & \textbf{0.79}             \\ \hline
		\end{tabular}
		\label{compare_metric}
	\end{center}
\end{table}

The proposed model surpassed both bicubic interpolation and the state-of-the-art SR model. Since the proposed model is benefiting from multi-scale gradient architecture, it can perform notably better in high-scale SR. However, in lower scales, the performance of the model is almost the same as other methods in terms of PSNR and SSIM as presented in Table \ref{4x}. 

\begin{table}[!htbp]
	\caption{Comparison of the performance of different prostate SR models for $4 \times$ SR.}
	
	\begin{center}
		\begin{tabular}{|c|c|c|}
			\hline
			\textbf{Model} & \textbf{PSNR} & \textbf{SSIM} \\ \hline
			SRResNet \cite{sood2018application}        & 21.03         & 0.70          \\ \hline
			SRGAN \cite{sood2019anisotropic}         & 21.27         & 0.66          \\ \hline
			Proposed Model & 21.09         & 0.74          \\ \hline
		\end{tabular}
		\label{4x}
	\end{center}
\end{table}

Moreover, because of the CapsNet in the discriminator, the geometrical relationships between features are more accurate in our model. Another way to compare the proposed system with the state-of-the-art is TSSA, as summarized in Table \ref{TSSU_table}.

\begin{table}[!htbp]
	\caption{Comparison of the performance of different prostate SR models.}
	
	\begin{center}
		\begin{tabular}{|c|c|c|c|}
			\hline
			\textbf{Model} & \textbf{Loss} & \textbf{Accuracy} & \textbf{TSSA} \\ \hline
			Classifier    & 3.73          & 86\%              & -             \\ \hline
			SRGAN \cite{sood2019anisotropic}         & 9.98          & 71\%              & 0.82          \\ \hline
			Proposed model & 7.25          & 79\%              & 0.88          \\ \hline
		\end{tabular}
		\label{TSSU_table}
	\end{center}
\end{table}

The results show that the proposed model achieved higher TSSA. It means that from a perspective of a deep learning-based architecture, our MSG-CapsGAN reconstructs the medical details more accurately. These results confirm that because of the robustness and the architecture of the proposed design, the medical details crucial for cancer diagnosis are reconstructed more accurately, with a notable margin.
 
As far as the complexity of the models is concerned, the number of parameters are compared in Table \ref{comp_param}. 

\begin{table}[!htbp]
	\caption{Number of parameters in the proposed model and the state-of-the-art work.}
	
	\begin{center}
		\begin{tabular}{|c|c|c|c|c|c|c|}
			\hline
			& \multicolumn{3}{c|}{\textbf{Generator}} & \multicolumn{3}{c|}{\textbf{Discriminator}} \\ \cline{2-7} 
			\textbf{Model} & trainable  & non-trainable  & total     & trainable   & non-trainable   & total       \\ \hline
			SRGAN \cite{sood2018application}          & 2,576,705  & 4,224          & 2,580,929 & 5,413,953   & 3,712           & 5,413,953   \\ \hline
			Proposed model & 927,619    & 1,445,696      & 2,373,315 & 3,676,065   & 144             & 3,676,209   \\ \hline
		\end{tabular}
		\label{comp_param}
	\end{center}
\end{table}

The total number of parameters in the generator of the proposed model is $8\%$ less than SRGAN. This difference is $63\%$ in trainable parameters. The proposed model is benefiting from non-trainable embedded CheXNet for feature exaction, as a result, notably less number of trainable parameters is required for efficient SR. The proposed discriminator has $32\%$ fewer parameters. CapsNet and MSG-GAN architecture are the main reasons for the drop in the number of parameters. CapsNet represents the features by a vector so it can outperform the CNN with less number of layers \cite{sabour2017dynamic}. MSG-GAN establishes many intermittent connections and proved the model with super-resolved images with all scales \cite{karnewar2020msg}. 

\section{Conclusion}
Prostate cancer is the second most common cancer worldwide. The ability to perform SR on prostate MRI can significantly increase the chance of early cancer diagnosis. As a result, the therapy can be started sooner and the chance of treatment success can be improved. In this paper, MSG-CapsGAN is used as the backbone of an effective SR MRI model. The feature extractor in this model is CheXNet. CheXNet is trained for lung disease detection so it can extract many informative and useful features from radiography scans. The input image is $28\times 28$ and the super-resolved output is $224\times 224$. The results are compared with the related works and the proposed model outperforms the state-of-the-art model with a margin greater than 1dB in PSNR and 0.1 and 0.15 in SSIM and MS-SSIM, respectively. Moreover, a new task-specific similarity assessment approach is introduced. This metric reflects the negative impact of the SR algorithm on the performance of a cancer detection system. It is demonstrated that while SRGAN can generate realistic outputs, it cannot preserve the medical details crucial for cancer detection as good as the proposed architecture. The robustness of the proposed model and the CapsNet in the discriminator are the main reasons behind the supremacy of the proposed architecture in reconstructing important medical details. 

The proposed model can be taken to the next level by employing ensemble learning paradigm. Different SR can be trained and another CNN combines all of the outputs and make the best super-resolved image. The CNN can be trained with the combination of GAN loss, perceptual loss, and TSSA, to make sure the output image is realistic, accurate, and useful for other DL models.

\bibliography{main.bib}
\bibliographystyle{IEEEtran}

\end{document}